\documentclass[runningheads]{llncs}
\usepackage{multirow}
\usepackage{booktabs}
\usepackage{graphicx}

\begin{document}
%
\title{BERT for Sentiment Analysis:\\Pre-trained and Fine-Tuned Alternatives}
\titlerunning{BERT for Sentiment Analysis: Pre-trained and Fine-Tuned Alternatives}
%

\author{Frederico Dias Souza\orcidID{0000-0002-4746-2136} \and João Baptista de Oliveira e Souza Filho\orcidID{0000-0001-6005-8480}}
\authorrunning{F. Souza and J. Souza Filho}
\institute{Eletrical Engineering Department, Federal University of Rio de Janeiro, Rio de Janeiro RJ 21941901, Brazil}

\maketitle 
\begin{abstract}
BERT has revolutionized the NLP field by enabling transfer learning with large language models that can capture complex textual patterns, reaching the state-of-the-art for an expressive number of NLP applications. For text classification tasks, BERT has already been extensively explored. However, aspects like how to better cope with the different embeddings provided by the BERT output layer and the usage of language-specific instead of multilingual models are not well studied in the literature, especially for the Brazilian Portuguese language. The purpose of this article is to conduct an extensive experimental study regarding different strategies for aggregating the features produced in the BERT output layer, with a focus on the sentiment analysis task. The experiments include BERT models trained with  Brazilian Portuguese corpora and the multilingual version, contemplating multiple aggregation strategies and open-source datasets with predefined training, validation, and test partitions to facilitate the reproducibility of the results. BERT achieved the highest ROC-AUC values for the majority of cases as compared to TF-IDF. Nonetheless, TF-IDF represents a good trade-off between the predictive performance and computational cost. 



\end{abstract}
\section{Introduction}
Text classification (TC) is one of the most widely studied natural language processing (NLP) tasks, exploring a range of methods. More recently, the Transformers architecture~\cite{transformers}, replacing the recurrence with the self-attention mechanism, enabled that large pre-trained language models could now be used to address several NLP tasks, leading to the  state-of-the-art in many of these applications~\cite{dl_tc_survey}.

BERT (Bidirectional Encoder Representations from Transformers)~\cite{bert} represents the most prominent Transformer-based model, extensively studied and evaluated in most NLP problems and benchmark datasets. However, more studies are required regarding pre-trained BERT approaches to identify the best manner of aggregating the multiple embeddings generated, especially for the Brazilian Portuguese language. 

Typically, classification models based on BERT embeddings use the output corresponding to the first token of the sequence (CLS). However, what if the multiple embeddings produced at this layer are combined to define a new document embedding? Would this procedure bring us some significant performance gain in the sentiment analysis scenario? To answer this question, we
analyzed three different BERT variants: the \emph{BERTimbau} Base and Large~\cite{bert}, a Portuguese BERT variant trained with the \emph{Brazilian Web as Corpus} (BrWaC)~\cite{brwac}, and the \emph{Multilingual BERT} (m-BERT)~\cite{bertgit}, trained on about 100 different languages, and considered a range of aggregation strategies over BERT outcomes to produce a single document embedding that is classified by a Logistic Regression (LR). 
The experiments included BERT models with and without fine-tuning, assuming different learning and dropout rates.

Another research question was how do these models generalize to other contextual settings? For this analysis, five open-source datasets with pre-defined training, test, and validation set partitions were considered in a  cross predictive performance experiment, i.e.,
embedding and/or classification models fine-tuned to a particular dataset were evaluated with  instances from  another one. 
In addition, a generalist model, developed with all datasets concatenated, was produced to infer if database specif models were significantly more accurate than a unique general embedding-classification model. To better position the results, the experiments also included the most classical embedding approach: the  \emph{term frequency-inverse document frequency} (TF-IDF)~\cite{first_tfidf} followed by a LR as a baseline. 

The main paper contribution is proposing different ways of using BERT for sentiment classification in Brazilian Portuguese texts. This analysis considered cost-benefit aspects, covering from more straightforward solutions to more computationally demanding approaches.

\section{Related Work}
\label{sec:related_work}
The most classical approach for text classification consists of extracting basic corpus statistics such as the word frequency or TF-IDF~\cite{first_tfidf} to generate large sparse embedding vectors with a size  equal to the vocabulary size. In these cases, Latent Semantic Analysis~\cite{lsa} may be useful for reducing the dimensionality of such vectors through the Singular Value Decomposition (SVD). As shown by Zhang et al.~\cite{zhang_cnn}, on some occasions, models using TF-IDF, despite being simpler and unable to capture complex text patterns, can achieve better results than more complex neural-based approaches.

Devlin et al.~\cite{bert} proposed the BERT (Bidirectional Encoder Representations from Transformers), one of the most popular Transformer-based architectures. Transfer-learning with BERT may consider two design options: pre-trained and fine-tuning. The pre-trained approach assumes BERT as a large fixed model for producing “unsupervised” features; therefore, only the model stacked over it is trained for a target application. Conversely, the fine-tuning strategy focus on updating BERT weights using labelled data from a specific task. 
Surprisingly, the authors presented competitive results in the named entity recognition (NER) task compared to the state-of-art just by exploiting the pre-trained approach. 

BERT's authors also open-sourced~\cite{bertgit} a variant with a multilingual purpose (m-BERT), trained on more than 100 languages, including Portuguese. In 2019, Souza et al.~\cite{bertimbau} open-sourced BERTimbau Base and Large, trained exclusively on Brazilian Portuguese corpora. This work focus on how to better explore m-BERT and BERTimbau in the sentiment analysis task.

Despite being a recent model, BERTimbau has already being applied to other tasks. Lopes et al.~\cite{Lopes_Correa_Freitas_2021} fine-tuned m-BERT and BERTimbau to an aspect extraction task, whereas Leite at al~\cite{toxic_leite} to toxic sentence classification, outperforming other bag-of-words solutions. Jiang et al.~\cite{irony_jiang} and Neto et al.~\cite{irony_neto} evaluated fine-tuning BERTimbau to an irony detection task. Carriço and Quaresma et al.~\cite{carrico_quaresma} exploited different ways of extracting features from its output layer (CLS token, vector maximum and vector average), considering a semantic similarity task.

Additionally, other authors considered producing large language models for the Brazilian Portuguese language. Paulo et al.~\cite{bertau} developed the BERTaú, a BERT Base variant trained with data from Itaú (the largest Latin American bank) virtual assistant, and reported better results than BERTimbau and m-BERT for the NER task. Carmo et al.~\cite{ptt5} open-sourced the PTT5 model, a T5 model trained on the BrWac corpus, the same used to train BERTimbau, achieving similar results to BERTimbau in a semantic similarity task.

\section{Datasets}

This work considered five user reviews datasets: \emph{Olist}~\cite{olist_2018}, \emph{B2W}~\cite{b2w_corpus}, \emph{Buscapé}~\cite{buscape_corpus}, \emph{UTLC-Apps}, and \emph{UTLC-Movies}~\cite{utlc_corpus}, available in a public repository~\cite{dataset_kaggle}, with predefined training, test, and validation set partitions. We also examined the results for all of them concatenated. In all evaluations, we considered the binary polarity target feature. Table \ref{tab:dataset_info} summarizes the number of samples, document length, vocabulary size, and the polarity distribution of each dataset.

\begin{table}[!htbp]
\caption{Summary of the datasets used in this work: number of samples, mean/median document length, vocabulary size, and the polarity  distribution.}
\centering
\begin{tabular}{c|c|c|c|c} 
\toprule
Dataset & \begin{tabular}[c]{@{}c@{}}Training/Validation/Test\\Samples\end{tabular} & \begin{tabular}[c]{@{}c@{}}Mean/Median\\Length\end{tabular} & \begin{tabular}[c]{@{}c@{}}Vocab size\\(1 gram)\end{tabular} & \begin{tabular}[c]{@{}c@{}}Labels\\Distribution\\(Positive)\end{tabular}\\ 
\midrule
Olist & 30k / 4k / 4k & 7 / 6 & 3.272 & 70.0\%\\
Buscapé & 59k / 7k / 7k & 25 / 17 & 13.470 & 90.8\%\\
B2W & 93k / 12k / 12k & 14 / 10 & 12.758 & 69.2\%\\
UTLC Apps & 775k / 97k / 97k & 7 / 5 & 28.283 & 77.5\%\\
UTLC Movies & 952k / 119k / 119k & 21 / 10 & 69.711 & 88.4\%\\
All & 1909k / 239k / 239k & 15 / 7 & 86.234 & 82.8\%\\
\bottomrule
\end{tabular}\label{tab:dataset_info}
\end{table}

\section{Models}
The experiments included two base models: m-BERT and BERTimbau, and two 
design strategies:  pre-trained and fine-tuned. The pre-trained solution considered just using BERT for producing document embeddings, subsequently fed to a LR model. The fine-tuned model considered adjusting the BERT weights to the sentiment analysis task. In both cases, the models adopted the \emph{AutoTokenizer} from \emph{Huggingface}~\cite{huggingface}, padding the sentences to 60 tokens and no extra processing was conducted over the documents.

The baseline adopted in this work was based on the results of Souza and Filho~\cite{ourpaper}, wherein TF-IDF and alternative word embedding strategies were evaluated in the same datasets and partitions. Therefore, this baseline considered the TF-IDF embedding followed by a LR, which performed best in all cases. The figure of merit was the ROC-AUC inferred over the test set.

\subsection{Pre-trained BERT}
\label{sec_emb_BERT}
Previous works~\cite{bert} reported that creative combinations of the token representations provided by the BERT outputs might lead to a significant performance improvement in the NER task, even without any fine-tuning of model parameters. Such findings strongly motivated the present work. 

The evaluated BERT models assumed documents constituted by one to the sixty tokens. To each token, this model produces a representation, having 768 or 1024 dimensions, in the case of Base and Large models, respectively, referred to as layers. In the following, we described the different approaches evaluated in this work to combine these embeddings. In parenthesis, the number of vector concatenations of each case is exhibited. Thus, the size of the vectors used for document representations  have from 768 (768$\times$1) up to 3072 (1024$\times$3) dimensions. 
\begin{enumerate}
\item \textbf{first} (1): layer corresponding to the first token, the [CLS] special token, created with the purpose of sentence classification, considered as  the default BERT embedding;
\item \textbf{second} (1): layer corresponding to the 2nd token;
\item \textbf{last} (1): layer corresponding to the last (60th) token;

\item \textbf{sum all} (1): sum of all 60 layers;
\item \textbf{mean all} (1): average of all 60 layers;
\item \textbf{sum all except first} (1): sum of all 59 layers, ignoring the first one;
\item \textbf{mean all except first} (1): average of all 59 layers, ignoring the first one;

\item \textbf{sum + first} (2): concatenation of the first layer with the sum of the remaining 59 layers;
\item \textbf{mean + first} (2): concatenation of the first layer with the average of the remaining 59 layers;

\item \textbf{first + mean + std} (3): concatenation of the first layer, the average and the standard deviation of the last 59 layers;
\item \textbf{first + mean + max} (3): concatenation of the first layer, the average and the maximum of the last 59 layers;
\item \textbf{mean + min + max} (3): concatenation of the average, minimum and maximum of the last 59 layers;
\item \textbf{quantiles 25, 50, 75} (3): concatenation of the quantiles 25\%, 50\%, and 75\% of the last 59 layers.

\end{enumerate}

For all these aggregations modalities, we also evaluated three different BERT models: BERTimbau Base and BERTimbau Large, trained exclusively with Brazilian Portuguese corpus, and m-BERT Base, trained in a multilingual schema.

\subsection{Fine-tuned BERT}

For BERT fine-tuning, a linear network was added on the top of the layer associated with the [CLS] token, targeting to predict one of the two sentiment classication classes, i.e., assuming as target-values 0 or 1. Network training adopted the Adam optimizer with weight decay, slanted triangular learning rates with a warm-up proportion of 0.1, and a maximum number of epochs equal to 4.

Design choices, such as the training and model hyperparameters, were based on Sun et al. ~\cite{finetune_bert}. Two base learning rates (2.5e-5 and 5e-5) and dropout rates (0 and 10\%) were evaluated considering the log-loss observed in the validation set. Experiments for hyperparameters' tuning restricted only to the \emph{Olist}, \emph{Buscapé}, and \emph{B2W},
due to the expressive computational efforts required for conducting this analysis in the significantly bigger remaining datasets. The best combination of the learning rate, dropout rate, and  number of epochs were 2.5e-5, 10\%, and one, for the BERTimbau models; whereas 2.5e-5, no dropout, and two, for the m-BERT. As expected, the m-BERT required one more training step to learn Portuguese language patterns.

\section{Results and Discussion}
\label{sec:results_discussion}

\subsection{Pre-trained BERT}

\begin{figure}[!htbp]
\centerline{\includegraphics[width=1.0\textwidth]{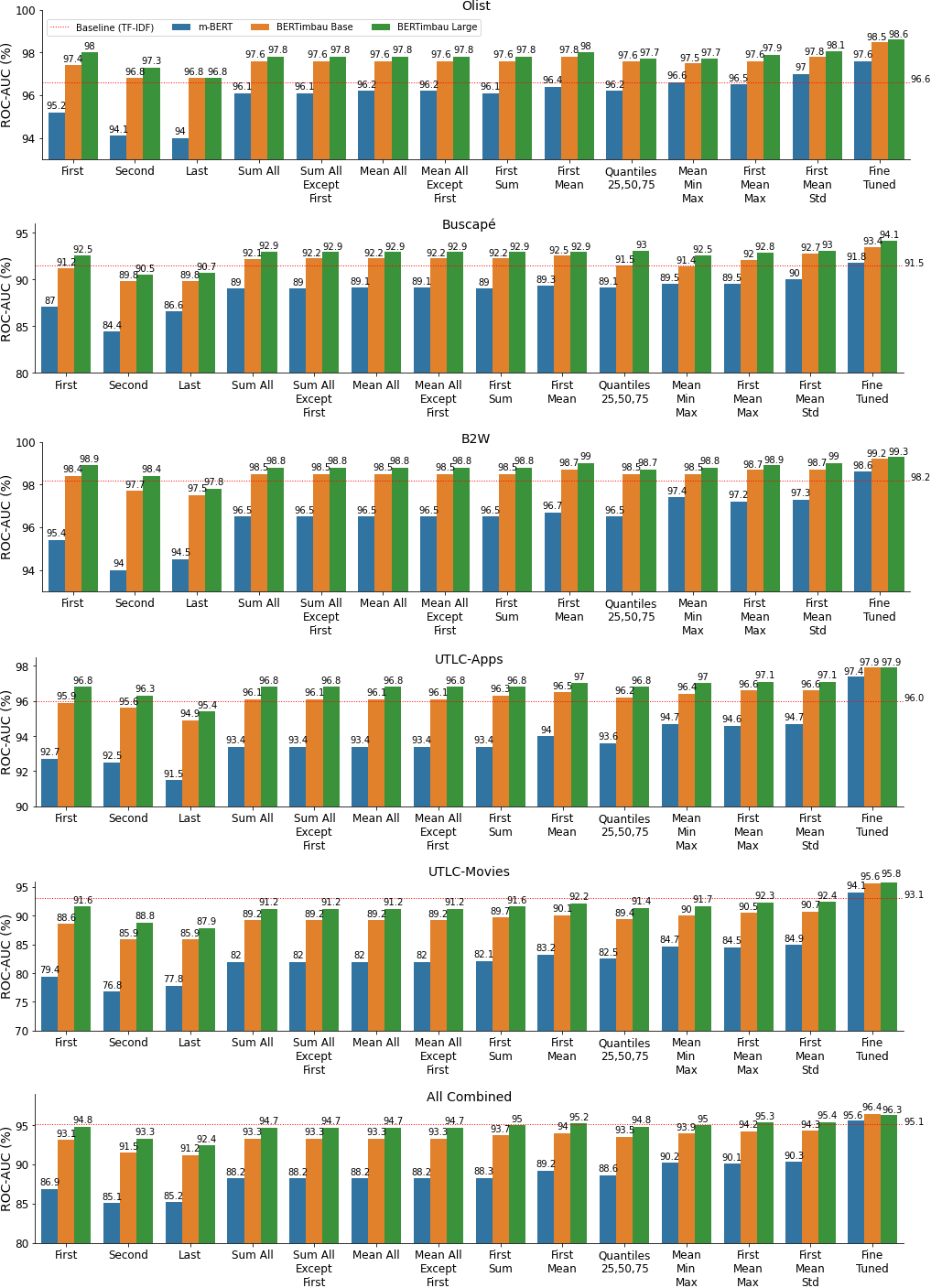}}
\caption{ROC-AUC (\%) per BERT Type, Embedding Type, and Dataset. The figure also includes the results for  fine-tuned BERT models. The horizontal line corresponds to the baseline model (TF-IDF with Logistic Regression).}
\label{fig:bert_results}
\end{figure}

Figure~\ref{fig:bert_results} compares the ROC-AUC values  considering different pre-trained BERT models and  aggregation strategies discussed in Section \ref{sec_emb_BERT}, including TF-IDF (horizontal red lines) and fine-tuned (to be discussed later).
For all datasets and aggregations methods, m-BERT Base is worse than BERTimbau Base, which is inferior to BERTimbau Large. In addition, except for  UTLC-Movies and  All-Combined datasets, these BERT embedding can outperform all previous approaches.

Table~\ref{tab:bert_rank} summarizes the average rankings restricted to the BERTimbau models for different datasets. 
The model Large performs better than the model Base in all cases. For both models, the best aggregations are: “first + mean + std”, “first + mean” and “first + mean + max”. 
These findings confirm that the first layer (CLS)  
carries, in fact, important information for classification, but such task can largely benefit from the remaining layers. Quantiles-based aggregations performed poorly, possibly due to not including the first layer content.

\begin{table}[!htbp]
\caption{Average Ranking (Avg. Rank) according to BERTimbau size and aggregation modality}
\centering
\begin{tabular}{ccccccccc} 
\cmidrule[\heavyrulewidth]{1-3}\cmidrule[\heavyrulewidth]{7-9}
\begin{tabular}[c]{@{}c@{}}BERT\\Size\end{tabular} & \begin{tabular}[c]{@{}c@{}}Aggregation\\Modality\end{tabular} & \begin{tabular}[c]{@{}c@{}}Avg.\\Rank\end{tabular} & & & & \begin{tabular}[c]{@{}c@{}}BERT\\Size\end{tabular} & \begin{tabular}[c]{@{}c@{}}Aggregation\\Modality\end{tabular} & \begin{tabular}[c]{@{}c@{}}Avg.\\Rank\end{tabular}\\
\cmidrule{1-3}\cmidrule{7-9}
Large & first + mean + std & 1.0 & & & & Base & first + mean + max & 13.8\\
Large & first + mean & 2.4 & & & & Base & first + sum & 15.0\\
Large & first + mean + max & 3.8 & & & & Base & sum all except 1st & 16.0\\
Large & first + sum & 4.6 & & & & Base & mean all except 1st & 16.0\\
Large & sum all except 1st & 5.2 & & & & Base & mean all & 16.0\\
Large & mean all except 1st & 5.2 & & & & Base & quantiles 25,50,75 & 16.8\\
Large & mean all & 5.2 & & & & Base & sum all & 16.8\\
Large & sum all & 5.2 & & & & Base & mean + min + max & 17.4\\
Large & first & 5.2 & & & & Large & second & 21.4\\
Large & mean + min + max & 7.0 & & & & Base & first & 22.4\\
Large & quantiles 25,50,75 & 7.2 & & & & Large & last & 24.0\\
Base & first + mean + std & 10.0 & & & & Base & second & 24.6\\
Base & first + mean & 11.0 & & & & Base & last & 25.2\\
\cmidrule[\heavyrulewidth]{1-3}\cmidrule[\heavyrulewidth]{7-9}
\end{tabular}\label{tab:bert_rank}
\end{table}

Previous results also bring us some important practical guidelines in operational scenarios with limited computational resources, for instance, those which restrictions over vRAM GPUs availability hinder the Large model usage: a simple aggregation of the type “first + mean + std” or even “first + mean” with the Base model can significantly boost its performance, 
reducing the gap between both. Additionally, all BERT embeddings evaluated here used numbers in 16 bits (dtype=“float16” in NumPy) precision, instead of 32 bits (the default). A quick analysis over the influence of using 16 and 32 bits in model training for BERTimbau Base and Large models with an aggregation of the type “first” has shown  
no significant impact on model performance. Nonetheless, the  reduction in the computational time and RAM usage with 16 bits of precision was impressive. Therefore, a useful practical lesson learned is to adopt numbers with 16 bits precision when dealing with pre-trained BERT models.

\subsection{Fine-tuned BERT}

Figure \ref{fig:bert_results} also exhibits the results for fine-tuned BERT models exclusively for the aggregation of the type “first + mean + std”. 
For all datasets, the fine-tuning allowed the new models to surpass all the pre-trained and baseline alternatives. The  \emph{UTLC-Movies} dataset 
was the one that most benefited from fine-tuning,
presenting relative gains for the ROC-AUC of 10\%, 4.9\%, and 3.4\%, considering m-BERT, BERTimbau Base, and BERTimbau Large, respectively. Moreover, for this dataset, BERT models only surpass 
the TF-IDF baseline by fine-tuning.
The BERT model that most benefited from fine-tuning was the m-BERT, despite not being able to reach the
BERTimbau performance after retraining. 
Therefore, the large dataset used to pre-train BERTimbau seems to contribute to its  significant predictive power.

\subsection{Cross-model comparison}
Cross-model comparisons were restricted to the BERTimbau Large models and “first + mean + std” embedding. Table~\ref{tab:bert_cross} summarizes the results. 

\begin{table}[!htbp]
\caption{Cross-comparison of ROC-AUC (\%) values for the BERTimbau pre-trained and fine-tuned models for different datasets (see text).}
\centering
\begin{tabular}{ccccccc} 
\toprule
\multicolumn{2}{c}{Model} & Olist & Buscapé & B2W & \begin{tabular}[c]{@{}c@{}}UTLC\\Apps\end{tabular} & \begin{tabular}[c]{@{}c@{}}UTLC\\Movies\end{tabular} \\ 
\midrule
\multirow{6}{*}{\begin{tabular}[c]{@{}c@{}}BERTimbau\\fine-tuned\\with\end{tabular}} & Olist & 97.9 & 93.4 & 99.1 & 97.4 & 93.2 \\
 & Buscapé & 98.3 & 93.9 & 99.0 & 97.3 & 93.3 \\
 & B2W & 98.2 & 93.4 & 99.2 & 97.5 & 93.7 \\
 & UTLC-Apps & 98.4 & 93.8 & 99.1 & 97.9 & 92.7 \\
 & UTLC-Movies & 98.2 & 93.7 & 99.1 & 97.4 & 95.9 \\
 & All-Combined & 98.4 & 93.7 & 99.0 & 97.5 & 95.3 \\ 
\midrule
\multicolumn{2}{c}{Pre-trained BERTimbau} & 98.1 & 93.0 & 99.0 & 97.1 & 92.4 \\
\bottomrule
\end{tabular}\label{tab:bert_cross}
\end{table}

Except for the Olist, the smallest dataset, specifically fine-tuning models to each dataset resulted in higher performance, as expected. 
The most extensive and complex content dataset, the \emph{UTLC-Movies}, exhibited the highest gains with retraining.  The general fine-tuned model obtained with the All-Combined dataset performed better than the pre-trained alternative but worse than the specialized models. Some datasets also seemed to benefit from fine-tuning, even if this procedure is conducted with another dataset. A notable example is again the \emph{UTLC-Movies}, when the retraining considered the  \emph{B2W} dataset, resulting in a gain of 1.3\% percentage points as compared to the pre-trained model. However, the gains observed with retraining to most datasets were inferior to 1\%, which might not be cost-effective for some applications.  

\section{Conclusion}
\label{sec:conclusion}
This work analyzed different ways of exploiting the BERT model for sentiment analysis on Brazilian Portuguese user reviews. An important finding is that the pre-trained BERTimbau embeddings performed much better than those related to the  multilingual m-BERT. Additionally, the proposed aggregation approach over  BERT outputs also brought  considerable gains over the more conventional scheme of selecting the first layer, i.e., the output related to the [CLS] token. 
For all datasets, except UTLC-Movies, some BERT embeddings configurations achieved the highest results among all evaluated models.
Nonetheless, the fine-tuning allowed BERT models to surpass all other alternatives, resulting in a significant m-BERT performance gain, despite it still performs worse than the BERTimbau model. Therefore, BERTimbau represents the best BERT variant for Brazilian Portuguese text classification tasks.

As following steps, we intend to evaluate the performance of other deep learning models, more complex than the TF-IDF baseline and simpler than BERT, such as CNNs and LSTMs. Furthermore, an interesting point for further studies is the fine-tuning schema, which may be modified to exploit better the large amount of annotated data available in the dataset considered in this work.

\bibliographystyle{paper}
\bibliography{paper}

\begin{thebibliography}{10}
\providecommand{\url}[1]{\texttt{#1}}
\providecommand{\urlprefix}{URL }
\providecommand{\doi}[1]{https://doi.org/#1}

\bibitem{ptt5}
Carmo, D., Piau, M., Campiotti, I., Nogueira, R., Lotufo, R.: \mbox{PTT5}:
  Pretraining and validating the \mbox{T5} model on \mbox{Brazilian Portuguese}
  data (2020)

\bibitem{carrico_quaresma}
Carrico, N., Quaresma, P.: Sentence embeddings and sentence similarity for
  \mbox{Portuguese} \mbox{FAQs}. In: IberSPEECH 2021. pp. 200--204 (03 2021)

\bibitem{bert}
Devlin, J., Chang, M.W., Lee, K., Toutanova, K.: {BERT}: Pre-training of deep
  bidirectional transformers for language understanding. In: Proceedings of the
  2019 Conference of the North {A}merican Chapter of the Association for
  Computational Linguistics: Human Language Technologies, Volume 1 (Long and
  Short Papers). pp. 4171--4186. Association for Computational Linguistics,
  Minneapolis, Minnesota (Jun 2019). \doi{10.18653/v1/N19-1423},
  \url{https://aclanthology.org/N19-1423}

\bibitem{bertau}
Finardi, P., Viegas, J.D., Ferreira, G.T., Mansano, A.F., Carid'a, V.F.:
  \mbox{BERTa{\'u}}: \mbox{Ita{\'u}} \mbox{BERT} for digital customer service.
  ArXiv  \textbf{abs/2101.12015} (2021)

\bibitem{bertgit}
Google: \mbox{BERT}. \url{https://github.com/google-research/bert} (2019)

\bibitem{buscape_corpus}
Hartmann, N., Avan{\c{c}}o, L., Balage, P., Duran, M., das Gra{\c{c}}as
  Volpe~Nunes, M., Pardo, T., Alu{\'\i}sio, S.: A large corpus of product
  reviews in {P}ortuguese: Tackling out-of-vocabulary words. In: Proceedings of
  the Ninth International Conference on Language Resources and Evaluation
  ({LREC}'14). European Language Resources Association (ELRA), Reykjavik,
  Iceland (May 2014)

\bibitem{irony_jiang}
Jiang, S., Chen, C., Lin, N., Chen, Z., Chen, J.: Irony detection in the
  \mbox{Portuguese} language using \mbox{BERT}. In: Iberian Languages
  Evaluation Forum 2021. pp. 891--897 (2021)

\bibitem{lsa}
Landauer, T.K., Foltz, P.W., Laham, D.: An introduction to latent semantic
  analysis. Discourse Processes  \textbf{25}(2-3),  259--284 (1998).
  \doi{10.1080/01638539809545028}

\bibitem{toxic_leite}
Leite, J.A., Silva, D.F., Bontcheva, K., Scarton, C.: Toxic language detection
  in social media for \mbox{Brazilian Portuguese}: \mbox{New} dataset and
  multilingual analysis. CoRR  \textbf{abs/2010.04543} (2020),
  \url{https://arxiv.org/abs/2010.04543}

\bibitem{Lopes_Correa_Freitas_2021}
Lopes, E., Correa, U., Freitas, L.: Exploring \mbox{BERT} for aspect extraction
  in \mbox{Portuguese} \mbox{Language}. The International FLAIRS Conference
  Proceedings  \textbf{34} (Apr 2021). \doi{10.32473/flairs.v34i1.128357},
  \url{https://journals.flvc.org/FLAIRS/article/view/128357}

\bibitem{dl_tc_survey}
Minaee, S., Kalchbrenner, N., Cambria, E., Nikzad, N., Chenaghlu, M., Gao, J.:
  Deep learning based text classification: \mbox{a} comprehensive review. arXiv
  preprint arXiv:2004.03705  (2020)

\bibitem{irony_neto}
Neto, A.M.S.A., Osti, B.A., de~Araujo~Azevedo, C., N{\'{o}}brega, F.A.A., Brum,
  H.B., Peinado, L.H.O., de~Menezes~Bittencourt, M., da~Silva, N.L.P.,
  C{\^{o}}rtes, P.O.: {SiDi-NLP-Team} at {IDPT2021:} {Irony} detection in
  {Portuguese} 2021. In: Montes, M., Rosso, P., Gonzalo, J., Arag{\'{o}}n,
  M.E., Agerri, R., Carmona, M.{\'{A}}.{\'{A}}., Mellado, E.{\'{A}}.,
  Carrillo{-}de{-}Albornoz, J., Chiruzzo, L., de~Freitas, L.A.,
  G{\'{o}}mez{-}Adorno, H., Guti{\'{e}}rrez, Y., Zafra, S.M.J., Lima, S., del
  Arco, F.M.P., Taul{\'{e}}, M. (eds.) Proceedings of the Iberian Languages
  Evaluation Forum (IberLEF 2021) co-located with the Conference of the Spanish
  Society for Natural Language Processing {(SEPLN} 2021), {XXXVII}
  International Conference of the Spanish Society for Natural Language
  Processing., M{\'{a}}laga, Spain, September, 2021. {CEUR} Workshop
  Proceedings, vol.~2943, pp. 933--939. CEUR-WS.org (2021),
  \url{http://ceur-ws.org/Vol-2943/idpt\_paper6.pdf}

\bibitem{olist_2018}
Olist: Brazilian e-commerce public dataset by \mbox{Olist} (Nov 2018),
  \url{https://www.kaggle.com/olistbr/brazilian-ecommerce}

\bibitem{b2w_corpus}
Real, L., Oshiro, M., Mafra, A.: \mbox{B2W-Reviews01} - an open product reviews
  corpus. STIL - Symposium in Information and Human Language Technology
  (2019), \url{https://github.com/b2wdigital/b2w-reviews01}

\bibitem{utlc_corpus}
Sousa, R.F.d., Brum, H.B., Nunes, M.d.G.V.: A bunch of helpfulness and
  sentiment corpora in \mbox{Brazilian Portuguese}. In: Symposium in
  Information and Human Language Technology - STIL. SBC (2019)

\bibitem{bertimbau}
Souza, F., Nogueira, R., Lotufo, R.: {BERT}imbau: pretrained {BERT} models for
  {B}razilian {P}ortuguese. In: 9th Brazilian Conference on Intelligent
  Systems, {BRACIS}, Rio Grande do Sul, Brazil, October 20-23 (to appear)
  (2020)

\bibitem{dataset_kaggle}
Souza, F.: {Brazilian} {Portuguese} sentiment analysis datasets (Jun 2021),
  \url{https://www.kaggle.com/fredericods/ptbr-sentiment-analysis-datasets}

\bibitem{ourpaper}
Souza, F., Filho, J.: Sentiment analysis on \mbox{Brazilian Portuguese} user
  reviews. In: IEEE Latin American Conference on Computational Intelligence
  2021 (preprint) (Dec 2021), \url{https://arxiv.org/abs/2112.05459}

\bibitem{first_tfidf}
Sparck~Jones, K.: A statistical interpretation of term specificity and its
  application in retrieval, p. 132–142. Taylor Graham Publishing, GBR (1988)

\bibitem{finetune_bert}
Sun, C., Qiu, X., Xu, Y., Huang, X.: How to fine-tune \mbox{BERT} for text
  classification? In: Sun, M., Huang, X., Ji, H., Liu, Z., Liu, Y. (eds.)
  Chinese Computational Linguistics. pp. 194--206. Springer International
  Publishing, Cham (2019)

\bibitem{transformers}
Vaswani, A., Shazeer, N., Parmar, N., Uszkoreit, J., Jones, L., Gomez, A.N.,
  Kaiser, L., Polosukhin, I.: Attention is all you need. In: Proceedings of the
  31st International Conference on Neural Information Processing Systems. p.
  6000–6010. NIPS'17, Curran Associates Inc., Red Hook, NY, USA (2017)

\bibitem{brwac}
Wagner~Filho, J.A., Wilkens, R., Idiart, M., Villavicencio, A.: The
  \mbox{BrWac} corpus: A new open resource for {B}razilian {P}ortuguese. In:
  Proceedings of the Eleventh International Conference on Language Resources
  and Evaluation ({LREC} 2018). European Language Resources Association (ELRA),
  Miyazaki, Japan (May 2018), \url{https://aclanthology.org/L18-1686}

\bibitem{huggingface}
Wolf, T., Debut, L., Sanh, V., Chaumond, J., Delangue, C., Moi, A., Cistac, P.,
  Rault, T., Louf, R., Funtowicz, M., Davison, J., Shleifer, S., von Platen,
  P., Ma, C., Jernite, Y., Plu, J., Xu, C., Scao, T.L., Gugger, S., Drame, M.,
  Lhoest, Q., Rush, A.M.: \mbox{HuggingFace's Transformers: State-of-the-art
  Natural Language Processing} (2020)

\bibitem{zhang_cnn}
Zhang, X., Zhao, J., LeCun, Y.: Character-level convolutional networks for text
  classification. In: Cortes, C., Lawrence, N., Lee, D., Sugiyama, M., Garnett,
  R. (eds.) Advances in Neural Information Processing Systems. vol.~28. Curran
  Associates, Inc. (2015),
  \url{https://proceedings.neurips.cc/paper/2015/file/250cf8b51c773f3f8dc8b4be867a9a02-Paper.pdf}

\end{thebibliography}

\end{document}